\documentclass[letterpaper]{article} 
\usepackage{aaai2027}  

\usepackage[hyphens]{url}  
\usepackage{graphicx} 
\usepackage{natbib}  
\usepackage{caption} 
\usepackage{algorithm}
\usepackage{algorithmic}

\usepackage{amsmath}
\usepackage{amssymb}

\usepackage{newfloat}
\usepackage{listings}
\DeclareCaptionStyle{ruled}{labelfont=normalfont,labelsep=colon,strut=off} 
\floatstyle{ruled}
\newfloat{listing}{tb}{lst}{}
\floatname{listing}{Listing}

\usepackage{booktabs}

\usepackage{multirow}
\usepackage[table,xcdraw]{xcolor}

\usepackage[latin1]{inputenc}
\usepackage{tikz}
\usetikzlibrary{shapes,arrows}
\usepackage{array}
\usepackage[tightpage]{preview}
\usetikzlibrary{arrows.meta, positioning}
\tikzset{
  box/.style = {
    rectangle, rounded corners, minimum width=3cm, minimum height=2cm,
    text centered, draw=gray, fill=white},
  arrow/.style={very thick, -Stealth},
  header/.style={
    label={[rectangle, fill=white, draw, anchor=center,
            minimum width=2cm, node font=\ttfamily,
            name=\tikzlastnode-header]north:{#1}}}
}
\PreviewEnvironment{tikzpicture}
\newcommand\blfootnote[1]{%
  \begingroup
  \renewcommand\thefootnote{}\footnote{#1}%
  \addtocounter{footnote}{-1}%
  \endgroup
}

\title{\textsc{DBLast}: Dependent Block Drafting for Stochastic Speculative Decoding}
\author {
    Amirmohammad Karimi,
    ~Chao Gao,
    ~Negar Hassanpour
}
\affiliations {
    Huawei Technologies Canada Co., Ltd. \\
    amirmohammad.karimi@h-partners.com, \{chao.gao4, negar.hassanpour2\}@huawei.com
}

\begin{document}
\nocopyright

\maketitle

\begin{abstract}
Speculative decoding accelerates large language models' inference by using a lightweight drafter to propose multiple future tokens and a target model to verify them. While recent block and diffusion-style drafters can predict several positions in a single pass, their training and sampling procedures are typically optimized for greedy decoding or assume that positions in the draft block are conditionally independent. This assumption becomes brittle in non-greedy speculative decoding, where the target distribution is deliberately stochastic and multiple continuations become plausible. We study this mismatch for block diffusion drafters and show that the accepted draft length degrades as the entropy of the target sampling distribution increases. We propose a dependent block drafter based on a low-rank latent mixture over token positions, complemented by an acceptance-oriented training objective that directly targets the expected verified length. Experiments with \hbox{Qwen3-4B} and \hbox{Qwen3-8B} on GSM8K, MT-Bench, HumanEval, and creative-writing benchmarks show that our approach, namely \textsc{DBLast}, consistently improves accepted length over independent block sampling, especially in higher-entropy decoding regimes.
\end{abstract}

\blfootnote{Preprint}
\section{Introduction}
\label{sec:intro}

Large language models (LLMs) are increasingly used in settings where inference cost, latency, and throughput are limiting factors. 
Speculative decoding~\citep{leviathan2023fast,chen2023accelerating} addresses this bottleneck by pairing a target model with a lightweight drafter. The drafter proposes future tokens, 
and the target model verifies those proposals in parallel while preserving the target distribution. 
Recent multi-token prediction~(MTP) drafters~\citep{gloeckle2024better,cai2024medusa} further 
reduce drafting overhead by predicting several future positions in a single low-cost forward pass. 
Block diffusion drafters such as DFlash~\citep{chen2026dflash} are a strong representative of this direction. Such methods use a non-autoregressive drafter to propose a block of future tokens and they can achieve high acceptance rates under greedy or low-entropy target decoding.

Greedy or low-entropy decoding, however, is not always desired.
Open-ended tasks such as dialogue and creative writing require diverse continuations~\citep{li2016diversity,holtzman2020curious,wiher2022decoding}, 
and reinforcement-learning post-training pipelines often rely on non-greedy rollouts for exploration rather than only high-probability completions~\citep{shao2024deepseekmath,liu2025understanding,yu2025dapo}.
When the target model is stochastic, achieving high acceptance rates requires distributional matching between the drafter and target model.
A common design in parallel drafting avoids the exponential cost of exact joint modeling by factorizing the proposal block across positions, as in MTP, Medusa-style, and DFlash-style drafters~\citep{gloeckle2024better,cai2024medusa,chen2026dflash}. The drafter is then trained to match the marginal target distribution at each future position, and at inference, it samples all block tokens independently given the prefix. 

Speculative verification, however, is sequential: the $i$th draft token is checked against the target distribution conditioned on the
prefix and previously accepted draft tokens.
Thus, a block with accurate per-position marginals can still be a poor speculative proposal if its tokens do not form a coherent conditional trajectory. 
The challenge is therefore to capture dependencies within the drafted block tokens, without sacrificing the one-pass parallel prediction that makes block speculative decoding efficient.

Existing approaches address this tension in two ways.
Sequential or semi-autoregressive draft heads~\citep{ankner2024hydra,li2024eagle,cheng2026dspark} model dependencies directly, but sacrifice part of the parallelism that makes block speculative decoding attractive.
Alternatively, tensor-decomposition and probabilistic-circuit approaches~\citep{basharin2025faster,grivas2025fast} define dependent joint distributions while retaining parallel prediction.
However, these methods primarily optimize the block negative log-likelihood~(NLL) of target tokens rather than the sequential acceptance process that determines speculative speedup.
Their empirical scope is also limited: \citeauthor{basharin2025faster}~focus mainly on smaller-scale or pretraining-style evaluations, while \citeauthor{grivas2025fast}~study byte-level language models.

To address these gaps, we propose to inject dependence into a DFlash-style block diffusion drafter for subword target models and use a training surrogate aligned with the verifier's accepted-prefix behavior. We refer to the resulting method as \textsc{DBlast}, a dependent block sampler trained with a loss on accepted length~(AL). 
\textsc{DBlast} augments the drafter with a categorical latent variable over the block by adding lightweight category output heads: 
conditioned on the latent category, all positions are still predicted in a single parallel pass, 
while marginalizing over categories induces a joint distribution that correlates positions within the block.

\begin{figure}[t]
    \centering
    \includegraphics[width=\linewidth]{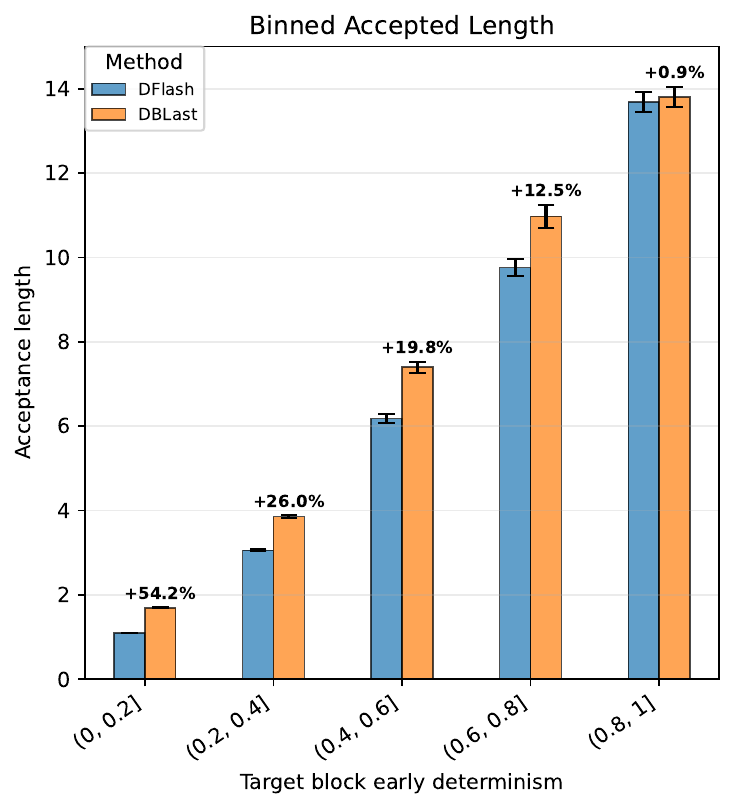}
    \caption{Accepted draft length vs. target-block early determinism on creative-writing continuations. We bin truncated points by the early determinism of the next 15-token target block. Bars report the mean accepted length, and error bars represent \(\pm 1\) standard error of the mean for the independent DFlash baseline and \textsc{DBlast}. Annotations show the relative improvement of \textsc{DBlast} over DFlash in each bin. Gains increase as target continuations become less deterministic. See Appendix Section \ref{sec:appendix-determinism} for the relationship between early determinism, target block diversity, and the diagnosis setting.}
    \label{fig:teaser_bar_plot}
\end{figure}

We further diagnose where independent block modeling fails. We finetune a \hbox{Qwen3-8B} DFlash checkpoint on task of creative writing in two cases of independent sampling with NLL and dependent sampling with AL loss training, then measure expected accepted length at randomly truncated target responses. We group samples by target-block early determinism, emphasizing early positions.%
\footnote{
\label{ft:determinism}
    For prefix $y$ and block length $b$, we define and empirically measure the target block early determinism as 
    \(\mathbb{E}_{x_{1:b} \sim p_{\mathrm{target}}(\cdot \mid y)}
    \left[\frac{1}{b}\sum_{j=1}^{b}\prod_{i=1}^{j}p_{\mathrm{target}}(x_i \mid y, x_{<i}) \right]\).
} 
As shown in Figure~\ref{fig:teaser_bar_plot}, independent proposals become increasingly fragile as determinism decreases. \textsc{DBlast} improves accepted length in every bin, with the largest relative gains in the least deterministic regions, supporting the value of coherent block-level alternatives and acceptance-oriented training for stochastic speculative decoding.

\paragraph{}
The following summarizes our contributions:
\begin{itemize}
    \item \textbf{Diagnosis.} We identify a mismatch between independent block sampling and non-greedy speculative verification, 
    and show empirically that the mismatch becomes more severe in less deterministic target-sampling regions, resulting in reduced accepted length.
    \item \textbf{Dependent Block Sampling.} We propose a dependent block drafter based on a canonical polyadic~(CP)-style latent mixture that preserves parallel token prediction while inducing correlations among positions. To our knowledge, this is the first application of CP-style dependent block proposals to subword block-diffusion drafters for non-greedy speculative sampling.
    \item \textbf{Loss on Accepted Length.} We propose an acceptance-oriented log-domain surrogate motivated by expected accepted length,
    with threshold-truncated prefixes for stable early training. This surrogate targets the verifier's sequential acceptance behavior, complementing dependent modeling when block likelihood alone does not optimize the accepted prefix length.
\end{itemize}

Our experimental results on \hbox{Qwen3-4B} and \hbox{Qwen3-8B} across GSM8K, \hbox{MT-Bench}, HumanEval, and creative-writing benchmarks show that \textsc{DBlast} consistently improves accepted length, with macro-average gains above $12\%$ in the high-entropy \hbox{Qwen3-8B} setting.

\section{Preliminaries}
\label{sec:Preliminaries}
\paragraph{Notation.}
Let $y$ denote the current prefix and let $x_{1:b}=(x_1,\ldots,x_b)$ denote a draft block of length $b$. We write $x_{<i}$ for the previously drafted tokens in the block. The target language model defines an autoregressive distribution $p$, and the drafter defines a proposal distribution $q$ over draft blocks. 
For any joint proposal $q(x_{1:b}\mid y)$ the conditional token-level proposal can be derived form the same joint proposal as:
\begin{equation}
    q(x_i\mid y,x_{<i})
    =
    \frac{q(x_{1:i}\mid y)}
         {q(x_{<i}\mid y)} .
    \label{eq:proposal-conditional}
\end{equation}

\paragraph{Speculative decoding.}
Given a sampled draft block $x_{1:b}$, non-greedy speculative verification accepts token $x_i$ with probability
\begin{equation}
    A_i
    =
    \min\!\left(
    1,\,
    \frac{p(x_i\mid y,x_{<i})}{q(x_i\mid y,x_{<i})}
    \right).
    \label{eq:accept-prob}
\end{equation}
The probability of accepting at least the first $i$ draft tokens is 
$
    \prod_{j=1}^{i}A_j,
$
therefore, the expected accepted length of the block is
\begin{equation}
    AL_b(x_{1:b})=\sum_{i=1}^{b}\prod_{j=1}^{i} A_j .
    \label{eq:acceptance-length}
\end{equation}

\paragraph{Block proposal families.}
A non-autoregressive DFlash-style drafter predicts the entire block at once in a single pass, with the standard proposal factorized across draft positions:

\begin{equation}
    q_{\mathrm{ind}}(x_{1:b}\mid y)
    =
    \prod_{i=1}^{b} q_i(x_i\mid y).
    \label{eq:independent-drafter}
\end{equation}
This proposal is efficient, but the joint probability of a block is only the product of its position-wise marginals.

We compare this baseline with a dependent block proposal based on a low-rank canonical polyadic mixture over block positions introduced by \citeauthor{basharin2025faster}~\citeyearpar{basharin2025faster} which defines the joint distribution using categorical latent variable $z\in\{1,\ldots,K\}$ as:
\begin{equation}
    q_{\mathrm{dep}}(x_{1:b}\mid y)
    =
    \sum_{z=1}^{K}
    q(z\mid y)
    \prod_{i=1}^{b} q_i(x_i\mid y,z).
    \label{eq:cp-drafter}
\end{equation}

\paragraph{Likelihood training baselines.}
The likelihood baselines minimize the negative log-likelihood of a target block $x_{1:b}^{\star}$ under the drafter proposal,
\begin{equation}
    \mathcal{L}_{\mathrm{NLL}}
    =
    -\log q(x_{1:b}^{\star}\mid y).
    \label{eq:block-nll}
\end{equation}

For the independent proposal in Equation~\ref{eq:independent-drafter}, this is the usual sum of per-position cross-entropies,

\begin{equation}
    \mathcal{L}_{\mathrm{NLL}}^{\mathrm{ind}}
    =
    -\sum_{i=1}^{b}\log q_i(x_i^{\star}\mid y).
    \label{eq:independent-nll}
\end{equation}

For the dependent proposal in Equation~\ref{eq:cp-drafter}, it is the marginalized block negative log-likelihood,

\begin{equation}
    \mathcal{L}_{\mathrm{NLL}}^{\mathrm{dep}}
    =
    -\log
    \sum_{z=1}^{K}
    q(z\mid y)
    \prod_{i=1}^{b}q_i(x_i^{\star}\mid y,z).
    \label{eq:dependent-nll}
\end{equation}

\vspace{1pt}
\section{Methodology}
\label{sec:method}
\textsc{DBlast} modifies a DFlash-style block diffusion drafter in two ways. First, it adds a categorical latent variable that induces dependencies among block positions while preserving parallel decoding. Second, it trains the drafter with an acceptance-oriented objective that indirectly rewards sampled blocks likely to survive speculative verification. We first describe the dependent block drafting and verification procedure in our framework, then explain the architectural modification, and finally present the training objective.

\subsection{Dependent Block Drafting and Verification}

We add dependencies with the latent-mixture proposal in Equation~\ref{eq:cp-drafter}. Sampling from Equation~\ref{eq:cp-drafter} is equivalently performed by first sampling $\hat{z}\sim q(z\mid y)$ and then sampling all block positions in parallel from $q_i(\cdot\mid y,\hat{z})$. The category acts as a block-level routing variable: each branch can represent a different continuation mode, while positions inside a selected branch are still generated in one parallel pass. Marginalizing over $z$ induces correlations among positions, and setting $K=1$ recovers the independent baseline.

\paragraph{Drafter sampling during inference.}
In our reported evaluations, we use a deterministic branch decoder: we sample the latent category from the learned prior, select the corresponding branch, and decode the branch tokens greedily within the DFlash block update. We model $q(z \mid y)$ by softmax over category logits, $\ell_z(y)$, and optionally apply an inference-time category temperature $Z_T$ before sampling the latent category,
\[
    q_{Z_T}(z\mid y)=\mathrm{softmax}(\ell_z(y)/Z_T).
\]
Smaller $Z_T$ sharpens the branch prior, while larger $Z_T$ increases category stochasticity. This greedy branch decoding results in block proposal distribution
\begin{equation}
q_{\mathrm{dep}}(x_{1:b}\mid y)
=
\sum_{z=1}^{K}
q_{Z_T}(z\mid y)
\prod_{i=1}^{b}\mathbf{1}\{x_i=\hat{x}_{i,z}(y)\},
\label{eq:greedy-branch-proposal}
\end{equation}
where
\[
    \hat{x}_{i,z}(y)=\arg\max_x q_i(x\mid y,z)
\]
defines the greedy token at position $i$ in branch $z$. 

\paragraph{Stochastic block verification.} For speculative verification of a sampled block in Equation~\ref{eq:accept-prob}, the token-level proposal conditional is obtained from the same joint distribution used in drafting by prefix marginalization:

\begin{equation}
    q_{\mathrm{dep}}(x_i\mid y,x_{<i})
    =
    \frac{q_{\mathrm{dep}}(x_{1:i}\mid y)}
         {q_{\mathrm{dep}}(x_{<i}\mid y)}.
    \label{eq:dep-conditional}
\end{equation}

Equivalently, after each accepted token, verification updates the category posterior,

\begin{equation}
    q(z\mid y,x_{<i})
    =
    \frac{
        q(z\mid y)\prod_{j<i}q_j(x_j\mid y,z)}
        {\sum_{z'=1}^{K}q(z'\mid y)\prod_{j<i}q_j(x_j\mid y,z')},
    \label{eq:category-posterior}
\end{equation}
and forms the drafter next-token distribution:
\begin{equation}
    q_{\mathrm{dep}}(\cdot\mid y,x_{<i})
    =
    \sum_{z=1}^{K}q(z\mid y,x_{<i})q_i(\cdot\mid y,z).
    \label{eq:dep-next-token-distribution}
\end{equation}

\paragraph{Distributional correctness.}
The non-greedy speculative-sampling proof applies to any proposal distribution $q$ as long as the acceptance probability and residual sampler use the same proposal conditional that generated the draft. Our change only modifies how the proposal block is parameterized and sampled. In our evaluations verification uses the conditional distribution induced by greedy branch decoding in Equation~\ref{eq:greedy-branch-proposal} and the replacement token after rejection is sampled from the usual normalized positive residual distribution $\max(p(\cdot\mid y,x_{<i})-q(\cdot\mid y,x_{<i}),0)$ using the corresponding proposal conditional in Equation~\ref{eq:dep-next-token-distribution}. Therefore, the vanilla speculative-sampling correction remains exact for the target distribution for the greedy-branch proposal used in our evaluations.

\paragraph{Architecture.}
We make a minimal output-side modification to the original DFlash architecture. The final transformer layer of DFlash trunk produces a hidden state $h_i$ for each block position. A category hidden expander maps this state to $K$ category-specific offsets, producing hidden branches

\begin{equation}
    h_{i,z}=h_i + g_z(h_i), \qquad z\in\{1,\ldots,K\}.
\end{equation}

\paragraph{}
The target LM head is then reused to obtain token logits from each branch to compute $q_i(.\mid y,z)$. A separate category-prior head maps the block-anchor hidden state to $K$ logits for \hbox{$q(z\mid y)$}. This modification adds only the expander and category-prior parameters. In this way the new model can be easily initialized from an existing DFlash checkpoint to benefit from the pre-trained knowledge. In the current experiments, the expander is a single linear layer and the prior is a single linear head; An example of parameter overhead for 4 categories is summarized in Table~\ref{tab:method-parameter-overhead}. As the expander can parallelize the computation across categories, all branch outputs can be computed in one model forward pass. We further investigate the latency overhead of this modification in Appendix Section \ref{sec:appendix-latency}. 

\begin{table}[t]
\centering
\renewcommand{\arraystretch}{1.4}
\begin{tabular}{@{}lrrr@{}}
\toprule
Target model & Hidden size & $K$ & Added parameters \\ \midrule
Qwen3-4B & 2560 & 4 & 26.2M ($+$5.2\%) \\
Qwen3-8B & 4096 & 4 & 67.1M ($+$6.7\%) \\ \bottomrule
\end{tabular}
\caption{Parameter overhead of the category hidden expander and category-prior head used in our experiments. For hidden size $H$, the added parameters are $KH^2+2KH+K$. Percentages in parentheses are relative to the corresponding original DFlash drafter checkpoints.
}
\label{tab:method-parameter-overhead}
\end{table}

\subsection{Acceptance-Oriented Training}

The NLL baselines in Equations~\ref{eq:block-nll}--\ref{eq:dependent-nll} are simple and stable, and they improve the likelihood of target-generated blocks under the proposal. However, they do not account for the verifier's sequential acceptance behavior. We therefore use expected accepted length to motivate an acceptance-oriented training surrogate. Decomposing accepted length into prefix-acceptance events gives
\begin{equation}
\begin{aligned}
\mathbb{E}_{x_{1:b} \sim q(\cdot\mid y)}
\!\left[AL_b(x_{1:b})\right]
&=
\sum_{\ell=1}^{b}
\mathbb{E}_{x_{1:\ell} \sim q(\cdot \mid y)}
\left[
    \prod_{j=1}^{\ell} A_j
\right] \\
&=
\sum_{\ell=1}^{b}
\mathbb{E}_{x_{1:\ell} \sim p(\cdot \mid y)}
\left[
    r_\ell
    \prod_{j=1}^{\ell} A_j
\right],
\end{aligned}
\label{eq:acceptance-length-objective}
\end{equation}
where
\[
    r_\ell=\frac{q(x_{1:\ell}\mid y)}{p(x_{1:\ell}\mid y)}
\]
is the importance ratio for prefix $x_{1:\ell}$. Here $p$ is the target distribution after applying the same temperature, top-$p$, and top-$k$ transformation used to generate the training trajectory and to compute the acceptance probabilities. The second equality in Equation~\ref{eq:acceptance-length-objective} changes measure separately for each accepted-prefix term. Ratios are evaluated only on target-sampled prefixes, for which $p(x_{1:\ell}\mid y)>0$.

This prefix-wise form also handles the zero probabilities introduced by top-$p$ and top-$k$ filtering. If a draft first leaves the filtered target support at position $j$, then according to \ref{eq:accept-prob}, $A_j=0$, so its contribution is zero for every prefix length $\ell\geq j$. Any positive contribution from the accepted prefix before position $j$ is retained in the separate terms with $\ell<j$. Thus, Equation~\ref{eq:acceptance-length-objective} does not require applying a full-block importance ratio to trajectories outside the filtered target support.

Equation~\ref{eq:acceptance-length-objective} lets us evaluate the exact accepted-length criterion using target-generated trajectories, but we do not directly optimize its raw importance-weighted terms. Early in training, the drafter and target distributions can have weak overlap, making these terms difficult to optimize. Instead, for each target prefix we define
\[
    S_\ell=\sum_{i=1}^{\ell}\prod_{j=1}^{i}A_j,
\]
where $S_\ell$ is the conditional expected accepted length within sub-block $x_{1:\ell}$. We retain prefixes through the first position at which $r_\ell$ falls below a threshold $\tau$. Formally,
\[
    \ell_\tau
    =
    \min\left(
        \{\ell\in\{1,\ldots,b\}:r_\ell<\tau\}
        \cup \{b\}
    \right),
\]
\[
    \qquad
    \mathcal{T}_\tau(x_{1:b})
    =
    \{1,\ldots,\ell_\tau\}.
\]
The resulting log-domain surrogate is
\begin{equation}
\begin{aligned}
\mathcal{J}_{\mathrm{AL}}
=
\mathbb{E}_{x_{1:b} \sim p(\cdot \mid y)}
\left[
\sum_{\ell\in\mathcal{T}_\tau(x_{1:b})}
    \left(
    \log r_\ell
    +
    \log S_\ell
    \right)
\right].
\end{aligned}
\label{eq:subblock-objective}
\end{equation}
The retained set depends on the current drafter and is treated as a stop-gradient selection rule: gradients pass through the selected terms but not through membership in $\mathcal{T}_\tau$. The logarithm and parameter-dependent truncation mean that Equation~\ref{eq:subblock-objective} is neither an unbiased estimator nor a claimed lower bound or statistically consistent estimator of expected accepted length. Rather, it is an acceptance-oriented surrogate that combines proposal probability, through $\log r_\ell$, with the verifier's accepted-prefix signal, through $\log S_\ell$, while excluding longer target prefixes to which the current drafter assigns negligible probability. Our training loss minimizes
\begin{equation}
    \mathcal{L}
    =
    - \mathcal{J}_{\mathrm{AL}} .
    \label{eq:training-loss}
\end{equation}
Likelihood-trained drafters are used as comparison baselines in our experiments. We test the practical effect of the surrogate in Section~\ref{sec:results}: threshold truncation focuses training on prefixes with non-negligible drafter--target overlap, while summing all retained sub-blocks in $\mathcal{T}_\tau(x_{1:b})$ provides denser supervision than optimizing a single retained prefix.

\paragraph{Target probabilities during training.}
The target probabilities in Equations~\ref{eq:acceptance-length-objective}--\ref{eq:subblock-objective} are computed from the target hidden states that are already used for the drafter. We apply the target LM head to the previous-position target hidden state, apply the same target sampling filter used to generate the target trajectory, and gather the probability assigned to each realized next token. Products of these per-token probabilities give $p(x_{1:\ell}\mid y)$ for each prefix length $\ell$.

\section{Experiments and Results}
\label{sec:results}

Experiments are designed to evaluate \textsc{DBlast} against baselines along four dimensions:
Section~\ref{sec:main_res} investigates wether dependent block proposals and acceptance-oriented training improve accepted length across a variety of tasks and target-sampling regimes. 
Section~\ref{sec:cat_num} studies how the number of latent categories $K$ affects performance in terms of accepted length. 
Sections~\ref{sec:stoch} and \ref{sec:acc_loss} examine the method's sensitivity to inference-time category temperature and acceptance-loss construction respectively. 

\begin{table*}[t]
\centering
\setlength{\tabcolsep}{1.5pt}
\renewcommand{\arraystretch}{1.4}

\newcommand{\gapA}{\rule{2pt}{0pt}}
\newcommand{\gapB}{\rule{6pt}{0pt}}

\resizebox{\textwidth}{!}{%
\begin{tabular}{
@{}
l
>{\raggedright\arraybackslash}p{2.25cm}
@{}c@{}
lll
@{}c@{}
lll
@{}
}
\toprule

\multirow{3}{*}{\textbf{Task}}
& \multirow{3}{*}{\textbf{Method}}
& \gapA
& \multicolumn{3}{c}{\textbf{Qwen3-4B}}
& \gapB
& \multicolumn{3}{c}{\textbf{Qwen3-8B}} \\

\cmidrule(lr){4-6}
\cmidrule(lr){8-10}

& &
\gapA
& \multicolumn{1}{c}{$T\!=\!0.7,\ p\!=\!0.8$}
& \multicolumn{1}{c}{$T\!=\!1.0,\ p\!=\!0.95$}
& \multicolumn{1}{c}{$T\!=\!1.5,\ p\!=\!0.95$}
& \gapB
& \multicolumn{1}{c}{$T\!=\!0.7,\ p\!=\!0.8$}
& \multicolumn{1}{c}{$T\!=\!1.0,\ p\!=\!0.95$}
& \multicolumn{1}{c}{$T\!=\!1.5,\ p\!=\!0.95$} \\

& &
\gapA
& \multicolumn{1}{c}{(high det.)}
& \multicolumn{1}{c}{(medium det.)}
& \multicolumn{1}{c}{(low det.)}
& \gapB
& \multicolumn{1}{c}{(high det.)}
& \multicolumn{1}{c}{(medium det.)}
& \multicolumn{1}{c}{(low det.)} \\

\midrule


\multirow{4}{*}{GSM8K}
& DFlash
& \gapA
& 5.84
& 5.48
& 4.98
& \gapB
& 5.77
& 5.31
& 4.71 \\

& DFlash + AL
& \gapA
& \underline{6.20 (+6.3\%)}
& \underline{5.82 (+6.2\%)}
& \underline{5.22 (+4.9\%)}
& \gapB
& \textbf{6.22 (+7.9\%)}
& \underline{5.66 (+6.5\%)}
& 5.01 (+6.4\%) \\

& DFlash + DS
& \gapA
& 5.96 (+2.0\%)
& 5.69 (+3.9\%)
& 5.16 (+3.8\%)
& \gapB
& \underline{5.90 (+2.4\%)}
& 5.60 (+5.5\%)
& \underline{5.04 (+7.0\%)} \\

& \textbf{\textsc{DBlast}}
& \gapA
& \textbf{6.29 (+7.7\%)}
& \textbf{6.04 (+10.3\%)}
& \textbf{5.53 (+11.1\%)}
& \gapB
& \textbf{6.22 (+7.8\%)}
& \textbf{5.90 (+11.2\%)}
& \textbf{5.35 (+13.7\%)} \\

\cmidrule(lr){1-10}


\multirow{4}{*}{MT-Bench}
& DFlash
& \gapA
& 2.98
& 2.85
& 2.69
& \gapB
& 2.85
& 2.69
& 2.42 \\

& DFlash + AL
& \gapA
& \textbf{3.14 (+5.3\%)}
& \underline{2.99 (+5.1\%)}
& \underline{2.78 (+3.4\%)}
& \gapB
& \textbf{2.99 (+4.8\%)}
& \underline{2.81 (+4.4\%)}
& 2.54 (+4.8\%) \\

& DFlash + DS
& \gapA
& 3.00 (+0.7\%)
& 2.93 (+3.0\%)
& 2.76 (+2.4\%)
& \gapB
& 2.88 (+0.9\%)
& 2.80 (+3.9\%)
& \underline{2.57 (+6.1\%)} \\

& \textbf{\textsc{DBlast}}
& \gapA
& \underline{3.10 (+3.8\%)}
& \textbf{3.04 (+6.8\%)}
& \textbf{2.88 (+6.8\%)}
& \gapB
& \underline{2.98 (+4.5\%)}
& \textbf{2.90 (+7.6\%)}
& \textbf{2.70 (+11.4\%)} \\

\cmidrule(lr){1-10}


\multirow{4}{*}{HumanEval }
& DFlash
& \gapA
& 4.84
& 4.61
& 4.36
& \gapB
& 5.03
& 4.56
& 3.97 \\

& DFlash + AL
& \gapA
& \underline{5.00 (+3.5\%)}
& \underline{4.82 (+4.4\%)}
& \underline{4.46 (+2.3\%)}
& \gapB
& \underline{5.15 (+2.4\%)}
& \underline{4.73 (+3.8\%)}
& 4.13 (+4.1\%) \\

& DFlash + DS
& \gapA
& 4.82 (-0.3\%)
& 4.74 (+2.7\%)
& \underline{4.46 (+2.4\%)}
& \gapB
& 4.99 (-0.9\%)
& 4.67 (+2.4\%)
& \underline{4.20 (+5.8\%)} \\

& \textbf{\textsc{DBlast}}
& \gapA
& \textbf{5.06 (+4.6\%)}
& \textbf{4.91 (+6.4\%)}
& \textbf{4.74 (+8.7\%)}
& \gapB
& \textbf{5.21 (+3.7\%)}
& \textbf{4.91 (+7.8\%)}
& \textbf{4.39 (+10.7\%)} \\

\cmidrule(lr){1-10}


\rowcolor[HTML]{D3D3D3}
&
DFlash
& \gapA
& 4.55
& 4.31
& 4.01
& \gapB
& 4.55
& 4.19
& 3.70 \\

\rowcolor[HTML]{D3D3D3}
&
DFlash + AL
& \gapA
& \underline{4.78 (+5.1\%)}
& \underline{4.54 (+5.3\%)}
& \underline{4.15 (+3.6\%)}
& \gapB
& \underline{4.79 (+5.2\%)}
& \underline{4.40 (+5.1\%)}
& 3.89 (+5.2\%) \\

\rowcolor[HTML]{D3D3D3}
&
DFlash + DS
& \gapA
& 4.59 (+0.9\%)
& 4.45 (+3.3\%)
& 4.13 (+3.0\%)
& \gapB
& 4.59 (+0.9\%)
& 4.36 (+4.0\%)
& \underline{3.93 (+6.4\%)} \\

\rowcolor[HTML]{D3D3D3}
\multirow{-4}{*}{Average}
& \textbf{\textsc{DBlast}}
& \gapA
& \textbf{4.81 (+5.8\%)}
& \textbf{4.67 (+8.2\%)}
& \textbf{4.38 (+9.3\%)}
& \gapB
& \textbf{4.81 (+5.6\%)}
& \textbf{4.57 (+9.2\%)}
& \textbf{4.15 (+12.1\%)} \\

\bottomrule
\end{tabular}%
}

\caption{
Holdout comparison of DFlash variants and \textsc{DBlast} using Qwen3-4B
and Qwen3-8B. Simple task names denote the corresponding held-out benchmark
splits. Results for both models are shown side by side at high, medium, and
low target determinism. The dependent-sampling rows use the $Z_T$ selected
for each model, loss, and target-sampling setting exclusively
on a disjoint $10\%$ split and frozen before evaluation on the remaining
$90\%$. Each cell reports
average accepted draft length rounded to two decimals, with percentage gain
over the corresponding DFlash baseline in parentheses. \textbf{Bold} marks
the largest accepted draft length, and \underline{underlined} marks the
second-largest distinct value within each model, task, and target-sampling
setting. Tied values receive the same formatting. Shaded rows report averages
across the three holdout benchmarks.
}
\label{tab:qwen3-model-comparison-ablation-holdout}
\end{table*}

\paragraph{Models.}
We evaluate drafters for \hbox{Qwen3-4B} and \hbox{Qwen3-8B} target models~\citep{qwen2025qwen3}. For each target model, we follow the architecture design in Section \ref{sec:method} and initialize the dependent drafter models from the official pre-trained DFlash checkpoints provided by~\citeauthor{chen2026dflash} with five full-attention layers. In our experiments we instantiate the same expander form for the $K\!=\!1$ baselines, so the capacity to generate a single branch is matched across objective comparisons.

\paragraph{Training.}
We use prompts from the Tulu3 SFT mixture~\citep{lambert2024tulu}. Training responses are generated by the target model with temperature $0.7$, top-$p$ $0.8$, and top-$k$ $20$.
Moreover, unless otherwise stated, AL denotes the sub-block objective in Equation~\ref{eq:subblock-objective} with threshold $\tau=0.1$. In all experiments we finetune the whole drafter model including the pre-trained DFlash parameters for one epoch. We adapt the SpecForge framework~\citep{specforge2025,li2026specforge} to our setting. Training hyperparameters are reported in Appendix Section \ref{sec:appendix-training}.

\paragraph{Tasks.}
For general evaluation, we use prompts from GSM8K~\citep{cobbe2021training}, MT-Bench~\citep{zheng2023judging}, and HumanEval~\citep{chen2021evaluating}. We evaluate three target-sampling settings: temperature $0.7$, top-$p$ $0.8$; temperature $1.0$, top-$p$ $0.95$; and temperature $1.5$, top-$p$ $0.95$, with top-$k$ fixed to $20$. 

\paragraph{Evaluation.}
We report macro-average accepted length while generating up to 256 tokens, excluding the bonus token sampled after accepted tokens. Verification uses vanilla non-greedy speculative decoding with the residual sampler. Independent $K\!=\!1$ drafters decode each block greedily. Dependent $K\!>\!1$ drafters first sample a category from the learned prior, optionally after applying category temperature $Z_T$, and then decode the selected branch greedily. For these greedy-branch evaluations, verification uses the conditional distribution induced by the deterministic mixture proposal in Equation~\ref{eq:greedy-branch-proposal}. Since the category prior controls branch diversity and was trained under a fixed target-sampling distribution, we treat $Z_T$ as an inference calibration parameter and report calibrated variants where applicable.

\paragraph{Category-temperature calibration.}
We partition the benchmark prompts into disjoint $10\%$ calibration and $90\%$ evaluation splits. We select $Z_T$ from $\{0.0,0.2,\ldots,1.4\}$ using only the calibration split, separately for each target model, training loss, and target-sampling setting, by maximizing macro-average accepted length across the three calibration tasks. We then freeze the selected value and report main results on the held-out evaluation split.
The evaluation sets contain 1,170 unique GSM8K prompts, 160 unique HumanEval prompts, and 72 unique MT-Bench prompts. We generate 1,170 continuations per task: one per GSM8K prompt and repeated generations from the HumanEval and MT-Bench prompt sets.

\subsection{Main results}
\label{sec:main_res}
Table~\ref{tab:qwen3-model-comparison-ablation-holdout} compares four drafter variants that isolate the two proposed ingredients: 
\textit{DFlash} denotes the independent \hbox{$K\!=\!1$} drafter trained with NLL; 
\textit{DFlash + AL} keeps \hbox{$K\!=\!1$} but trains with AL; 
\textit{DFlash + DS} uses dependent \hbox{$K\!=\!4$} sampling trained with NLL; and 
\textsc{DBlast} combines dependent \hbox{$K\!=\!4$} sampling trained with AL. 
These variants are evaluated across a variety of tasks, target models, and target-sampling settings.

The full ablation separates the effect of the training objective from that of dependency modeling. Acceptance-oriented training improves the independent $K\!=\!1$ drafter over NLL optimization strategy in all average rows, while dependent sampling is most useful as target sampling becomes less deterministic. Combining the two gives the strongest average accepted length for both target models in all three target-sampling regimes. For \hbox{Qwen3-8B}, the average gain of \textsc{DBlast} over DFlash grows from $5.6\%$ in the high-determinism setting to $12.1\%$ in the low-determinism setting. In the same low-determinism setting, independent AL improves by $5.1\%$ and dependent NLL improves by $6.4\%$, showing that dependency and acceptance-oriented training are complementary.

\subsection{Ablation on number of categories}
\label{sec:cat_num}

Table~\ref{tab:qwen3-4b-category-ablation-t1p5} studies the number of latent categories using the \hbox{Qwen3-4B} DFlash checkpoint in the high-entropy target setting. We fine-tune for one epoch on Tulu3-generated responses and evaluate the mean accepted length across GSM8K, HumanEval, and MT-Bench. Increasing $K$ improves both NLL and AL variants, but most of the benefit is reached by $K\!=\!4$. AL outperforms NLL for every $K$, and its gain from increasing $K$ is larger, suggesting that the acceptance-oriented objective uses the additional branch capacity more effectively.
\begin{table}[t]
\centering
\renewcommand{\arraystretch}{1.}
\begin{tabular}{@{}lrrrr@{}}
\toprule
Training loss
  & $K\!=\!1$
  & $K\!=\!2$
  & $K\!=\!4$
  & $K\!=\!8$ \\ \midrule
NLL
  & $4.00 $
  & $4.10 $
  & $4.10 $
  & $4.13 $ \\
AL
  & $4.17 $
  & $4.31 $
  & $4.39 $
  & $4.41 $ \\ \bottomrule
\end{tabular}
\caption{Qwen3-4B category-count ablation at target sampling
$T\!=\!1.5$, $p\!=\!0.95$ and 1 epoch of finetuning. Entries report
mean accepted draft length across GSM8K, HumanEval, and MT-Bench.}
\label{tab:qwen3-4b-category-ablation-t1p5}
\end{table}


\subsection{Ablation on drafter stochasticity at inference}
\label{sec:stoch}
\begin{figure}[t]
    \centering
    \includegraphics[width=1.0\linewidth]{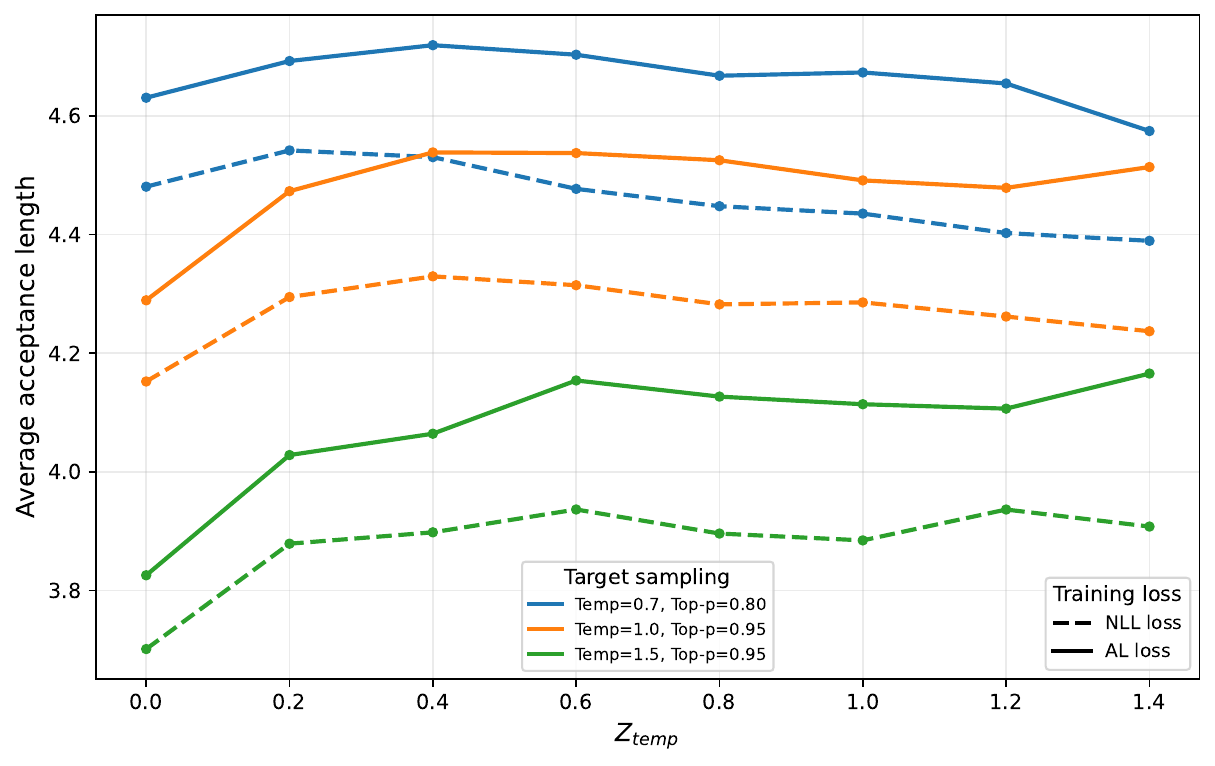}
    \caption{Average Acceptance Length for Dependent DFlash-Qwen3-8B with $K\!=\!4$ under NLL and AL variants over category temperatures. Colors correspond to target sampling parameters.}
\vspace{15pt}
    \label{fig:z_temp_ablation}
\end{figure}
Figure~\ref{fig:z_temp_ablation} evaluates inference-time category temperature for the dependent $K\!=\!4$ \hbox{Qwen3-8B} drafter under NLL and AL training. Across the full range of $Z_T$, the AL-trained drafter consistently remains above the NLL-trained drafter in terms of average accepted length. In both cases, the best category temperature also shifts upward as the target distribution becomes less deterministic. This argmax trend suggests a meaningful match between target stochasticity and draft stochasticity: broader target distributions benefit from more category diversity in the drafter. Very large $Z_T$ does not keep improving accepted length, so the benefit comes from calibrated diversity rather than arbitrary noise.

\subsection{Ablation on acceptance-oriented training loss}
\label{sec:acc_loss}
\begin{table}[t]
\centering
\renewcommand{\arraystretch}{1.}
\begin{tabular}{@{}lrrr@{}}
\toprule
Num Categories
  & Whole block
  & One prefix
  & All prefixes \\ \midrule
$K\!=\!1$
  & $4.73 $
  & $4.87 $
  & $4.99 $ \\
$K\!=\!4$
  & $4.98 $
  & $5.18 $
  & $5.33 $ \\ \bottomrule
\end{tabular}
\caption{Qwen3-8B ablation of the acceptance-oriented log surrogate at target sampling
$T\!=\!1.5$, $p\!=\!0.95$ and 1 epoch of finetuning. Entries report mean accepted draft length on GSM8K.}
\vspace{15pt}
\label{tab:qwen3-8b-loss-ablation}
\end{table}

Table~\ref{tab:qwen3-8b-loss-ablation} ablates the AL construction using 140K-step \hbox{Qwen3-8B} checkpoints evaluated on GSM8K in the high-entropy target setting with $Z_T=1.0$. We compare three variants of the log-domain surrogate: 
(i)~using only the whole block~($\ell=b$), 
(ii)~optimizing one threshold-selected prefix, and 
(iii)~summing all retained prefixes with $\tau=0.1$. 
The whole-block log surrogate is close to the NLL baseline in Table~\ref{tab:qwen3-model-comparison-ablation-holdout}. This variant optimizes $\log r_b+\log S_b$, rather than the raw importance-weighted quantity in Equation~\ref{eq:acceptance-length-objective}. When the drafter is still poor, $r_b=q(x_{1:b}\mid y)/p(x_{1:b}\mid y)$ is often very small, while the acceptance factors $A_j$ saturate near $1$ on target-sampled prefixes. Consequently, $S_b$ changes slowly and the gradient is dominated by $\log r_b$, making the update resemble block NLL. Truncating low-overlap prefixes improves the training signal, and summing all retained prefixes gives the strongest result for both $K\!=\!1$ and $K\!=\!4$.

\section{Related Work}
\label{sec:related}

\paragraph{Speculative decoding and speculative sampling.}
Speculative decoding accelerates autoregressive generation by proposing tokens with a cheap draft model and verifying them with the target model \citep{leviathan2023fast}. Speculative sampling extends the same principle to stochastic decoding while preserving the target distribution \citep{chen2023accelerating}. Most analyses and systems focus on improving draft accuracy or target-model verification efficiency. Our work focuses on a different failure mode: when the drafter proposes an entire block non-autoregressively, the block's joint structure can matter as much as the marginal accuracy of individual positions.

\paragraph{Parallel and block drafters.}
Multi-token prediction and block drafting reduce the number of target-model invocations by proposing several future tokens per pass~\citep{gloeckle2024better,cai2024medusa}. DFlash uses a block diffusion drafter to generate draft tokens efficiently and is especially effective for greedy speculative decoding~\citep{chen2026dflash}. D-PACE changes the training objective for parallel speculative drafters by assigning dynamic position-aware cross-entropy weights~\citep{wu2026dpace}. Other draft-head methods introduce sequential or semi-autoregressive structure inside the draft, including Hydra, EAGLE, and DSpark~\citep{ankner2024hydra,li2024eagle,cheng2026dspark}. These methods improve drafting quality, but they either retain independent block sampling or introduce within-draft serialization. Our focus is the conditional mismatch introduced by independent non-autoregressive block sampling under non-greedy verification.

\paragraph{Dependent multi-token modeling.}
Recent tensor-decomposition and probabilistic-circuit approaches to multi-token prediction show that explicitly modeling dependencies among future positions can improve the expressiveness of parallel predictors~\citep{basharin2025faster,grivas2025fast}. These methods typically train by maximizing the likelihood of a target future block under the proposal distribution. For a factorized proposal, this reduces to the standard sum of per-position cross-entropies; for a CP-style latent-mixture proposal, it becomes a marginalized block negative log-likelihood that distributes credit across latent branches according to the learned prior. We use these objectives as likelihood baselines for both independent and dependent drafters.

\section{Limitations}
\label{sec:limitations}

The current study establishes the draft proposal quality benefits of both dependent block drafting and acceptance-oriented training, while leaving three directions for further investigation.

\paragraph{Effect of training data stochasticity}
We primarily evaluated the drafters trained with a set of target responses that are generated using the default sampling parameters. However, inherently, dependent modeling may benefit from more stochasticity in training target blocks since the drafter is exposed to a more accurate representation of target block distribution. We leave the exploration of training dataset stochasticity to future work.

\paragraph{Acceptance-oriented objective.}
We optimize a threshold-truncated, log-domain surrogate motivated by expected accepted length.
The surrogate is designed to provide a practical training signal when drafter--target overlap is limited,
and our experiments show that it consistently improves accepted length.
It does not,
however,
provide the formal guarantees of an unbiased estimator or lower bound,
leaving tighter acceptance-aligned objectives as an interesting direction for future work.

\paragraph{Training and inference proposals.}
Training uses the differentiable soft latent-mixture proposal,
whereas our primary evaluation samples a category and greedily decodes its corresponding branch.
This design enables differentiable training while retaining efficient block decoding at inference,
and its consistent gains across models,
tasks,
and sampling regimes demonstrate effective empirical transfer.
A formal characterization of the relationship between the soft training distribution and the resulting greedy-branch proposal remains open.
\section{Conclusion}
\label{sec:conclusion}
We studied block diffusion drafters for non-greedy speculative decoding and showed that independent block sampling creates a fundamental mismatch in non-greedy speculative decoding:
the drafter predicts future positions separately,
while the target verifies them conditionally in a sequential manner.
We show that this mismatch becomes increasingly costly as target decoding grows more stochastic,
causing independently sampled blocks to lose acceptance precisely when multiple continuations are plausible.

To address this limitation, we introduced \textsc{DBlast}, a dependent block drafter that represents coherent block-level alternatives through a low-rank latent mixture and trains them with an acceptance-oriented objective aligned with sequential verification.
\textsc{DBlast} preserves efficient one-pass parallel drafting while consistently improving accepted length across math, coding, chat, and creative-writing tasks,
with a $12.1\%$ macro-average gain for \hbox{Qwen3-8B} in the highest-entropy setting.
These results demonstrate that within-block dependency modeling and acceptance-aligned training are complementary ingredients for efficient stochastic speculative decoding.

\bibliography{refs}
\clearpage
\appendix
\providecommand{\placeholder}[1]{\textbf{[PLACEHOLDER: #1]}}

\section{Supplementary Material}
\label{sec:appendix}

This supplement provides the technical and experimental details supporting
the main paper. Section~\ref{sec:appendix-correctness} provides DBLast overall design and specifies the proposal
used by the reported greedy-branch decoder and its exact verification.
Section~\ref{sec:appendix-determinism} documents the early-determinism
diagnosis. Section \ref{sec:appendix-latency} investigates the latency overhead of adding dependency to the drafter, Section \ref{sec:appendix-training} gives the full training configuration, and finally
Section \ref{sec:appendix-qualitative}
covers qualitative behavior of DBLast.

\section{DBLast Drafting and Verification}
\label{sec:appendix-correctness}

\begin{figure}[t]
\centering
\includegraphics[width=\linewidth]{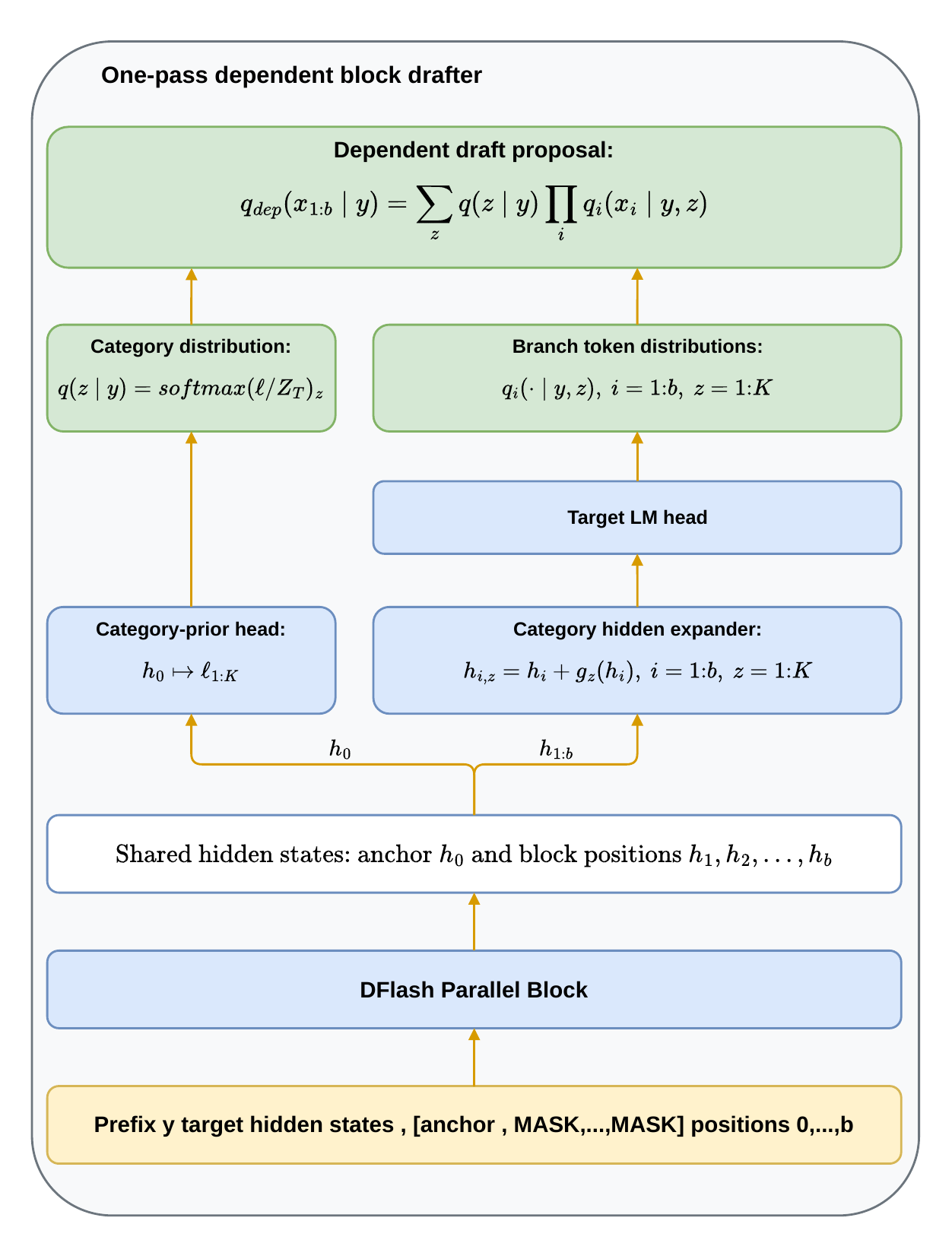}
\caption{DBLast architecture.}
\label{fig:appendix-architecture}
\end{figure}

Figure \ref{fig:appendix-architecture} illustrates the overall implementation of low-rank mixture dependency modeling in DBLast. Inputs are target prefix hidden states, anchor and masked tokens. The DFlash backbone computes the shared block hidden states; the category-prior head defines the category distribution; and the hidden expander computes all category-specific hidden states in parallel. The category distribution and the branch token distributions define the dependent draft proposal.

The latent-mixture proposal is
\begin{equation}
q(x_{1:b}\mid y)
=\sum_{z=1}^{K}q(z \mid y)\prod_{i=1}^{b}q_i(x_i\mid y,z).
\label{eq:appendix-soft-proposal}
\end{equation}
which is differentiable and used in our training to comupute the draft probabilty of a sampled (sub-)block.

\subsection{Greedy-Branch Proposal}

The reported experiments sample a category and decode its branch greedily.
Let $\widehat{x}_{i,z}(y)=\arg\max_v q_i(v\mid y,z)$. For $Z_T>0$, define
$\pi_z^{(Z_T)}(y)=\operatorname{softmax}(\ell(y)/Z_T)_z$. The inference
proposal is therefore the finite mixture of deterministic blocks
\begin{equation}
q_{\mathrm{greedy}}(x_{1:b}\mid y)
=\sum_{z=1}^{K}\pi_z^{(Z_T)}(y)
\prod_{i=1}^{b}\mathbf{1}\{x_i=\widehat{x}_{i,z}(y)\}.
\label{eq:appendix-greedy-proposal}
\end{equation}
For $Z_T=0$, the highest-prior category is selected. Algorithm \ref{alg:appendix-dependent-verification} summarizes drafting and verification at inference.

\subsection{Prefix-Marginalized Conditional}

For draft prefix $x_{<i}$, let
$\mathcal{C}_i(x_{<i})=\{z:\widehat{x}_{<i,z}(y)=x_{<i}\}$ be the set of
consistent branches. Their normalized posterior weights are
\begin{equation}
\widetilde{\pi}_{i,z}
=\frac{\pi_z^{(Z_T)}(y)\mathbf{1}\{z\in\mathcal{C}_i\}}
{\sum_{z'\in\mathcal{C}_i}\pi_{z'}^{(Z_T)}(y)}.
\end{equation}
The conditional proposal used by verification is
\begin{equation}
q_{\mathrm{greedy}}(v\mid y,x_{<i})
=\sum_{z\in\mathcal{C}_i}\widetilde{\pi}_{i,z}
\mathbf{1}\{v=\widehat{x}_{i,z}(y)\}.
\label{eq:appendix-greedy-conditional}
\end{equation}
This sparse distribution has support on at most $K$ tokens and naturally
handles branches that share an initial prefix. Every reported experiment uses Equation~\ref{eq:appendix-greedy-conditional}, rather than the soft token conditionals in Equation~\ref{eq:appendix-soft-proposal}, for acceptance and residual sampling.

\subsection{Distributional Correctness}

At position $i$, the drafted token is accepted with probability
\begin{equation}
\alpha_i=\min\left\{1,
\frac{p(x_i\mid y,x_{<i})}
{q_{\mathrm{greedy}}(x_i\mid y,x_{<i})}\right\}.
\end{equation}
After a rejection, the replacement is drawn from the normalized positive
residual $[p(\cdot\mid y,x_{<i})-
q_{\mathrm{greedy}}(\cdot\mid y,x_{<i})]_+$. The mass emitted through
acceptance is $\min\{p(v),q_{\mathrm{greedy}}(v)\}$, and the residual supplies
exactly the remaining target mass. Applying this argument after each accepted
prefix preserves the target distribution for the complete output sequence.

\begin{algorithm}[t]
\caption{Greedy-Branch Dependent Speculative Verification}
\label{alg:appendix-dependent-verification}
\begin{algorithmic}[1]
\STATE Given prefix $y$, compute category logits and all $K$ greedy branches.
\STATE Sample $z\sim\pi^{(Z_T)}(y)$ and select
$x_{1:b}=\widehat{x}_{1:b,z}(y)$.
\STATE Run the target on the draft to obtain
$p_i(\cdot)=p(\cdot\mid y,x_{<i})$.
\FOR{$i=1,\ldots,b$}
  \STATE Compute $q_{\mathrm{greedy}}(\cdot\mid y,x_{<i})$ using
  Equation~\ref{eq:appendix-greedy-conditional}.
  \STATE Accept $x_i$ with probability
  $\min\{1,p_i(x_i)/q_{\mathrm{greedy}}(x_i\mid y,x_{<i})\}$.
  \IF{$x_i$ is rejected}
    \STATE Sample from normalized
    $[p_i(\cdot)-q_{\mathrm{greedy}}(\cdot\mid y,x_{<i})]_+$ and stop.
  \ENDIF
\ENDFOR
\STATE If all tokens are accepted, sample the usual target bonus token.
\end{algorithmic}
\end{algorithm}

\section{Target-Block Early Determinism}
\label{sec:appendix-determinism}

For prefix $y$ and block length $b$, we define target-block early determinism as
\begin{equation}
\mathcal{D}_b(y)=\mathbb{E}_{x_{1:b}\sim p_{\mathrm{target}}(\cdot\mid y)}
\left[\frac{1}{b}\sum_{j=1}^{b}\prod_{i=1}^{j}
p_{\mathrm{target}}(x_i\mid y,x_{<i})\right].
\label{eq:appendix-early-determinism}
\end{equation}
For fixed $j$,
\begin{equation}
\mathbb{E}_{X_{1:j}\sim p_{\mathrm{target}}}
[p_{\mathrm{target}}(X_{1:j}\mid y)]
=\sum_{u\in\mathcal{V}^{j}}p_{\mathrm{target}}(u\mid y)^2.
\end{equation}
Thus, $\mathcal{D}_b(y)$ averages the collision probabilities of the first
$1,\ldots,b$ target-prefix distributions. It is large when probability mass
is concentrated on a few continuations and small when many branches are
plausible. 

Using $M$ independently sampled blocks, we estimate it as
\begin{equation}
\widehat{\mathcal{D}}_b(y)=\frac{1}{Mb}\sum_{m=1}^{M}\sum_{j=1}^{b}
\prod_{i=1}^{j} p_{\mathrm{target}}
(x_i^{(m)}\mid y,x_{<i}^{(m)}).
\label{eq:appendix-early-determinism-estimator}
\end{equation}

\subsection{Creative-Writing Diagnostic}
The diagnostic uses Qwen3-8B and compares an independent $K=1$ DFlash
drafter trained with NLL against a dependent $K=4$ drafter trained with AL.
Creative-writing prompts are taken from the EQ-Bench Creative Writing
Benchmark v3~\citep{creative-writing-bench-v3}, specifically the
\texttt{creative\_writing\_prompts\_v3.json} prompt file with 330 unique prompts.
We train drafter variants using target responses with sampling parameters of temperature $1.0$, top-$p$ $0.95$, and top-$k$ $20$. One hundred training responses are generated per prompt, and the drafters are fine-tuned on the resulting dataset for 150K steps.

To analyze the trained drafters, we uniformly sampled 30k truncations from a set of target generated responses with same prompts in training. We then measured the expected acceptance length for one block drafting and verification at each truncation. Figure~1 in the main paper shows data corresponding to early-determinism-estimator in Equation~\ref{eq:appendix-early-determinism-estimator}, with $b=15$, $M=20$, and target sampling parameters 
temperature $1.0$, top-$p$ $0.95$, and top-$k$ $20$.
Table~\ref{tab:appendix-creative-bin-counts} reports the number of evaluated samples in each target-block early-determinism bin. The significant mass of samples with the lowest determinism (first bin ) and the gain reported in Figure 1 of the paper confirm the importance of dependency modeling and acceptance-oriented training in tasks such as creative writing.

\begin{table}[t]
\centering
\small
\begin{tabular}{@{}lr@{}}
\toprule
Early-determinism bin & Number of samples \\
\midrule
$(0.0,0.2]$ & 24,078 \\
$(0.2,0.4]$ & 5,104 \\
$(0.4,0.6]$ & 614 \\
$(0.6,0.8]$ & 145 \\
$(0.8,1.0]$ & 59 \\
\midrule
Total & 30,000 \\
\bottomrule
\end{tabular}
\caption{Creative-writing sample counts in the target-block
early-determinism bins used for Figure~1 of the main paper. All methods are
evaluated on the same sampled truncation points, so the bin counts are shared
across methods.}
\label{tab:appendix-creative-bin-counts}
\end{table}

\section{Latency Overhead}
\label{sec:appendix-latency}

Accepted draft length isolates proposal quality, but DBLast additionally
computes category-conditioned branches and the prefix-marginalized mixture
used during verification. To quantify the practical cost of these operations,
we measure the wall-clock latency of one complete speculative iteration---one
drafting step followed by one target-verification step---using Qwen3-4B target model on a consumer level GPU. Each
reported value is averaged over all speculative iterations executed while
generating 1,000 responses, with each response capped at 256 new tokens at batch size of 1. All
methods use the same prompts, target model, decoding configuration, and
implementation. Calculated overhead is relative to independent DFlash without category expander head.

\begin{table}[t]
\centering
\small
\begin{tabular}{@{}lrrr@{}}
\toprule
Method & $K$ & Latency (ms) & Overhead \\
\midrule
DFlash (no expander) & 1 & 65.3 & -- \\
DFlash (with expander) & 1 & 65.5 & $+0.3\%$ \\
\textsc{DBLast} & 2 & 66.0 & $+1.1\%$ \\
\textsc{DBLast} & 4 & 66.2 & $+1.4\%$ \\
\textsc{DBLast} & 8 & 67.5 & $+3.4\%$ \\
\bottomrule
\end{tabular}
\caption{Average wall-clock latency per speculative iteration for Qwen3-4B.}
\label{tab:appendix-latency}
\end{table}

\begin{table}[t]
\centering
\small
\begin{tabular}{@{}lrrr@{}}
\toprule
Method & $K$ & Latency (ms) & Overhead \\
\midrule
DFlash (no expander) & 1 & 103.8 & -- \\
DFlash (with expander) & 1 & 105.2 & $+1.3\%$ \\
\textsc{DBLast} & 4 & 105.5 & $+1.6\%$ \\
\bottomrule
\end{tabular}
\caption{Average wall-clock latency per speculative iteration for Qwen3-8B.}
\label{tab:appendix-latency-qwen3-8b}
\end{table}

\begin{table*}[ht]
\centering
\small
\begin{tabular}{@{}p{0.27\textwidth}p{0.67\textwidth}@{}}
\toprule
Item & Value \\
\midrule
Target models & Qwen3-4B and Qwen3-8B \\
Drafter initialization & Official DFlash checkpoints; five full-attention
layers \\
Training prompts & Tulu3 SFT mixture \\
Target sampling & Temperature $0.7$, top-$p$ $0.8$, top-$k$ $20$ \\
Sequence length & 2048 \\
Draft block length & 15 \\
Categories & Main $K=4$; ablation $K\in\{1,2,4,8\}$ \\
AL threshold & $\tau=0.1$ unless otherwise stated \\
Optimization & Learning rate $10^{-4}$; warmup ratio $0.04$; cosine decay;
gradient max-norm $1.0$ \\
Duration & One epoch; 170k steps for Qwen3-8B and 190k steps for Qwern3-4B main runs unless otherwise stated \\
Optimizer & Adam \\
Batching & micro-batch=4, accumulation=1, global batch size = 4 \\
Software & SpecForge adaptation \\
\bottomrule
\end{tabular}
\caption{Training configuration.}
\label{tab:appendix-training-config}
\end{table*}

As shown in Table~\ref{tab:appendix-latency}, the main $K=4$ DBLast
configuration increases per-iteration latency from 65.3 ms to 66.2 ms, an
overhead of 0.9 ms (1.4\%). The overhead remains modest as the number of
categories grows, reaching 3.4\% at $K=8$. Thus, in this setting, the
additional category expander and mixture calculations have a small effect on the combined drafting-and-verification cost. 

\section{Training Configuration}
\label{sec:appendix-training}

The target models are Qwen3-4B and Qwen3-8B. Drafters are initialized from
the official DFlash checkpoints with five full-attention layers. Unless
otherwise stated, the entire drafter is fine-tuned for one epoch on target generated responses
from the Tulu3 SFT mixture. Training responses use temperature $0.7$, top-$p$
$0.8$, and top-$k$ $20$. Table~\ref{tab:appendix-training-config} provides the full training configuration.

\section{Qualitative Examples}
\label{sec:appendix-qualitative}

Table~\ref{tab:creative-qualitative} illustrates the proposal distributions
at the first draft step for a creative-writing prompt. The target samples
exhibit several plausible opening modes: some begin with the morning light or
air, while others begin with the sun over the streets or the Colosseum. The
independent DFlash drafter can expose only one block for this context.

In contrast, \textsc{DBLast} assigns non-negligible prior mass to all four
categories ($0.20$--$0.31$) and exposes distinct and coherent block-level alternatives in
the same parallel drafting pass.

\begin{table*}[t]
\centering
\footnotesize
\setlength{\tabcolsep}{4pt}
\begin{tabular}{p{0.19\textwidth}p{0.75\textwidth}}
\toprule
Source & Continuations or draft candidates for the same creative-writing prompt \\
\midrule
Target samples &
\begin{tabular}[t]{@{}p{0.75\textwidth}@{}}
\textit{1.} ``The morning light filtered through the narrow window of my cell, casting long shadows'' \\
\textit{2.} ``The morning air was thick with the scent of damp earth and the acrid'' \\
\textit{3.} ``The morning air was thick with the scent of sweat, smoke, and the'' \\
\textit{4.} ``The sun was high, casting a golden sheen over the dusty streets of'' \\
\textit{5.} ``The sun had not yet risen when I slipped through the iron gates of the'' \\
\textit{6.} ``The sun had not yet risen over the Colosseum, its marble'' \\
\textit{7.} ``The sun had long since climbed above the rooftops of the Colosse'' \\
\textit{8.} ``The morning air was thick with the scent of damp earth and the tang of'' \\
\textit{9.} ``The sun was high when I rose, the kind of sun that beats down'' \\
\textit{10.} ``The morning air was thick with the scent of damp earth and the faint tang''
\end{tabular} \\
\midrule
DFlash, NLL   
&
\begin{tabular}[t]{@{}p{0.12\textwidth}p{0.60\textwidth}@{}}
$q(z{=}1)=1.00$ & ``The morning air was thick the the scent of the,, the,,''
\end{tabular} \\
\midrule
\textsc{DBlast}, $K=4$ &
\begin{tabular}[t]{@{}p{0.12\textwidth}p{0.60\textwidth}@{}}
$q(z{=}1)=0.20$ & ``The morning air was thick with the scent of burning earth and the lingering tang'' \\
$q(z{=}2)=0.31$ & ``The sun had barely begun to the horizon of of olive and the, oil'' \\
$q(z{=}3)=0.28$ & ``The sun was high over the Colosseum, and the lingering of'' \\
$q(z{=}4)=0.20$ & ``The morning sun hadpt the edge, of sl,, and the the''
\end{tabular} \\
\bottomrule
\end{tabular}
\caption{
    Example from the first draft step of a creative-writing run. The target row shows the first ten target-sampled continuations. Non-greedy target sampling admits multiple plausible continuations for the prompt, while independent block drafting exposes a single candidate. The dependent drafter exposes multiple latent-category candidates and learned prior probabilities.
}
\label{tab:creative-qualitative}
\end{table*}

\end{document}